\pdfoutput=1

\documentclass[11pt]{article}
\usepackage{amssymb}
\usepackage[preprint]{acl}

\usepackage{times}
\usepackage{latexsym}
\usepackage{url}            
\usepackage{booktabs}       
\usepackage{amsfonts}       
\usepackage{nicefrac}       
\usepackage{microtype}      
\usepackage{latexsym}
\usepackage{amssymb}
\usepackage{amsmath}
\usepackage{booktabs}
\usepackage{enumerate}
\usepackage{graphicx}
\usepackage{subfigure}
\usepackage{xspace}
\usepackage{float}
\usepackage{bbm}
\usepackage{bm}
\usepackage{multirow}
\usepackage{booktabs}
\usepackage{color}
\usepackage{framed}
\usepackage{stfloats}
\usepackage{iitem}
\usepackage{amssymb}
\usepackage{pifont}
\usepackage{arydshln}
\usepackage{enumitem}
\usepackage{wrapfig}
\usepackage{algorithm}
\usepackage{algpseudocode}
\usepackage{array}
\usepackage{tabularx}
\usepackage{booktabs}
\usepackage{float}

\usepackage[T1]{fontenc}
\usepackage{colortbl}
\definecolor{maroon}{cmyk}{0,0.87,0.68,0.32}

\usepackage[utf8]{inputenc}

\usepackage{microtype}

\definecolor{bittersweet}{rgb}{1.0, 0.44, 0.37}
\definecolor{mygreen}{rgb}{0.29, 0.7, 0.48}
\usepackage{pifont}
\definecolor{demphcolor}{RGB}{144,144,144}

\definecolor{mygray}{gray}{0.4}
\definecolor{autopurple}{HTML}{7030A0}
\definecolor{dyna_yellow}{HTML}{BF9000}
\definecolor{adaptive_blue}{HTML}{0070C0}
\definecolor{darksalmon}{rgb}{0.91, 0.59, 0.48}
\definecolor{emerald}{rgb}{0.31, 0.78, 0.47}
\definecolor{green(pigment)}{rgb}{0.0, 0.65, 0.31}
\definecolor{amaranth}{rgb}{0.9, 0.17, 0.31}
\definecolor{iris}{rgb}{0.35, 0.31, 0.81}
\definecolor{uu}{rgb}{0.95, 0.51, 0.51}
\definecolor{spirodiscoball}{rgb}{0.06, 0.75, 0.99}
\usepackage{svg}

\usepackage{cleveref}

\usepackage{xspace}

\definecolor{ada_blue}{rgb}{0,205,205}
\definecolor{glt_red}{rgb}{109,205,255}
\definecolor{MorandiBlue}{RGB}{118,134,146}

\definecolor{demphcolor}{RGB}{144,144,144}
\definecolor{mygray}{gray}{0.4}
\definecolor{autopurple}{HTML}{7030A0}
\definecolor{dyna_yellow}{HTML}{BF9000}
\definecolor{adaptive_blue}{HTML}{0070C0}
\definecolor{darkgrey}{RGB}{120,120,120}
\definecolor{mygrey}{RGB}{200,200,200}

\usepackage{makecell}
\usepackage{tabulary}
\usepackage[most]{tcolorbox}

\definecolor{myblue}{HTML}{00CDCD}
\definecolor{champagne}{rgb}{0.74, 0.83, 0.9}
\definecolor{champagne}{rgb}{0.97, 0.91, 0.81}

\usepackage{inconsolata}
\usepackage{multirow}
\newcommand{\method}{SeaPO\xspace}

\title{\method: \underline{S}trategic \underline{E}rror \underline{A}mplification for \\ Robust \underline{P}reference \underline{O}ptimization of Large Language Models}

\author{
Jun Rao$^{1}$,
 Yunjie Liao$^{1}$,~
 Xuebo Liu$^{1}$\thanks{~Corresponding Author},
 Zepeng Lin$^{1}$,~\\
 \bf{Lian Lian$^{2}$},
 \bf{Dong Jin$^{2}$}, \bf{Shengjun Cheng}$^{2}$,~
  \bf{Jun Yu}$^{3}$, and
    \bf{Min Zhang}$^{1}$\\
    \textsuperscript{\rm1}Institute of Computing and Intelligence, Harbin Institute of Technology, Shenzhen \\
    \textsuperscript{\rm2}Huawei Cloud Computing Technologies Co., Ltd. \\
    \textsuperscript{\rm3}School of Intelligence Science and Engineering, Harbin Institute of Technology, Shenzhen \\
    \texttt{\{rao7jun,yunjie445,zepenglin11\}@gmail.com}\\
    \texttt{\{liuxuebo,yujun,zhangmin2021\}@hit.edu.cn}\\
    \texttt{\{lianlian3,jindong2,chengshengjun\}@huawei.com}
}

\begin{document}
\maketitle
\begin{abstract}

Existing alignment methods for preference optimization of large language models (LLMs) aim to enhance model performance by utilizing pairs of positive and negative samples. However, due to the limited capacity of models in scoring or generating responses, the quality of positive and negative samples may become similar during training, which complicates optimization for preference learning. 
To address this issue, we introduce SeaPO, a {\bf S}trategic {\bf E}rror {\bf A}mplification method that leverages three error types commonly occurring in LLMs to introduce specific error patterns into the model {\bf P}reference {\bf O}ptimization. 
This strategy ensures that negative samples are more erroneous than positive samples and preference-based training is employed to mitigate the occurrence of these errors, thereby enhancing model performance. Evaluations across five capability dimensions and different model scales (1.5B to 14B) demonstrate that the generated data significantly improved overall model performance, particularly in terms of truthfulness, with improvements of 5–10 percentage points observed. Further analysis reveals that task performance varies depending on the error types introduced. Injecting the most common error types improves performance in related tasks, while a mix of error types leads to a broader performance enhancement: most tasks show stable improvements, while a few tasks exhibit significant gains.

\end{abstract}

\section{Introduction}
\begin{figure}[t]
    \centering
    \includegraphics[width=1.0\linewidth]{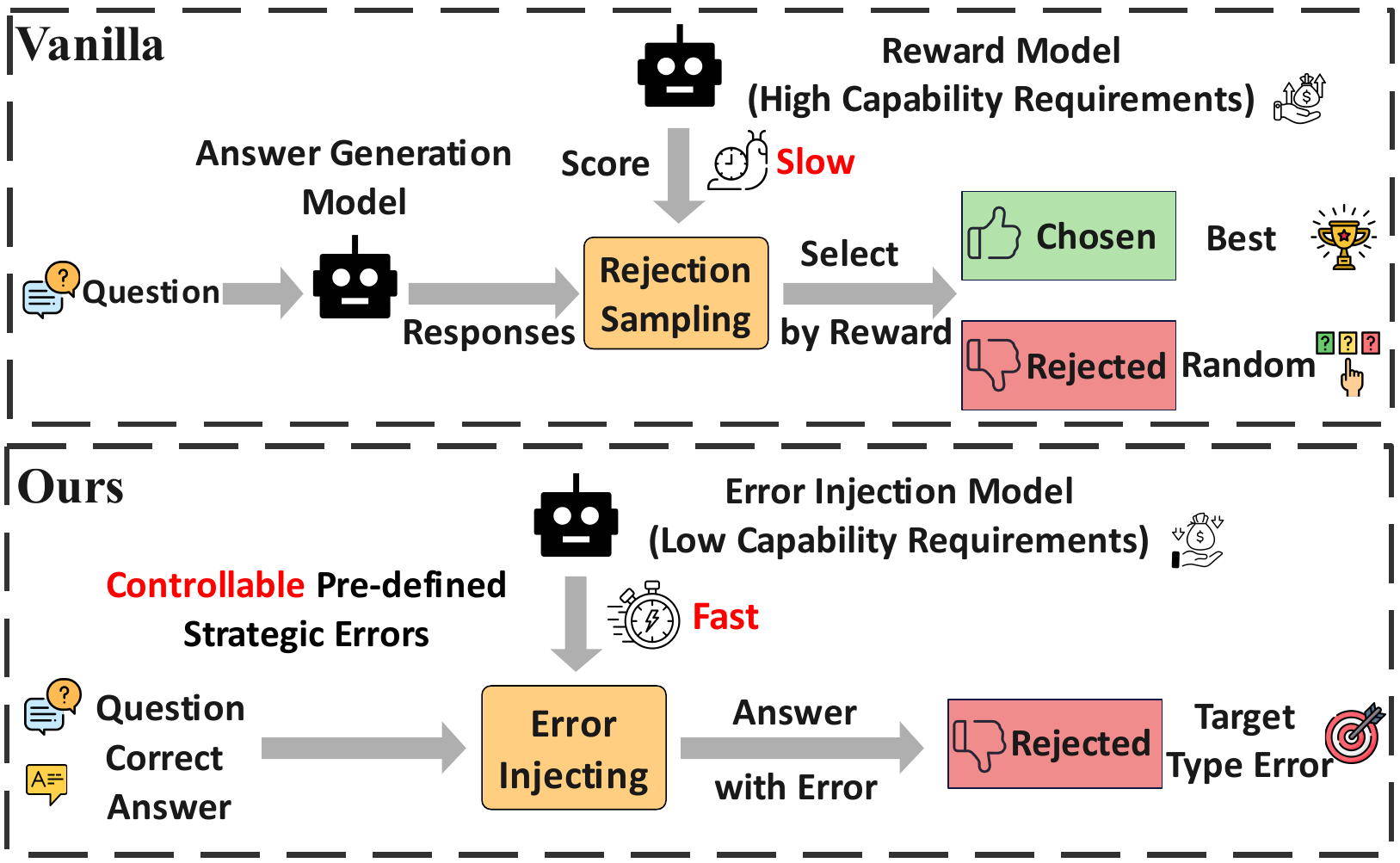}
    \caption{Due to the limitations of the reward model and the sampling, the rejection may be as good as the chosen one, making preference-based optimization more difficult. In the vanilla approach, positive and negative samples are obtained through multiple sampling and a scoring model, which are required to identify positive and negative samples. In contrast, our method only requires introducing specific error types to the original responses without the need for rejection sampling. }
    \label{intro}
\end{figure}

Since the advent of GPT4~\cite{gpt4}, data have become the major affect factor in large language models (LLMs) of capability~\cite{wang2023far,qi2023limitation,NewTermBenchmarking_DJL}. 
Extensive research has explored various aspects of data, including knowledge distillation~\cite{alpaca-gpt4,rao2023DCD,pesf-kd,bao2023post,yang2024learning}, generation~\cite{zhang2025featuresdeserveattentiongraphguided,yang2025doublingdataminutesultrafast},  filtering~\cite{jiang2023mistral,rao-etal-2025-apt}, selection~\cite{anonymous2024alpagasus,xiao2024feature}, and training~\cite{commonIT}. 
Past work primarily focused on optimizing samples in the supervised fine-tuning phase, i.e., improving the quality~\cite{longpre2023flan,cot,xie2023data,anonymous2024alpagasus,llama,llama3}, diversity~\cite{xu2023baize} and length~\cite{zhao2024long} of the training samples.
Recent work has also explored the effect of data in DPO~\cite{dpo}, using methods such as rejection sampling~\cite{yuan2023scalingrelationshiplearningmathematical,Cui2024UltraFeedbackBL}, iteration of negative samples~\cite{chen2024selfplayfinetuningconvertsweak,dong2024selfboostinglargelanguagemodels} (which progressively improves the quality of negative samples), and applying constraints to prevent overfitting the model on existing data~\cite{gpo}, including length constraints~\cite{meng2024simpo,han-etal-2024-length} and regularization constraints~\cite{pal2024smaug}.

However, due to the inherent limitations of a model's capabilities~\cite{jiang2023mistral,qwen,deepseekv2}, achieving human-like precision in specific tasks~\cite{zhao2024seer}, such as evaluating sentence quality or generating accurate responses~\cite{wei-etal-2025-chain,han2025attributestextualgenesleveraging} to complex questions, remains challenging. 
This results in scoring models potentially misjudging~\cite{zhao2024medico,zhao2024funnelrag} positive and negative samples, leading to minimal differences between them and, consequently, difficulties in preference-based training, which can result in potential logical or hallucination generation.

To address this challenge, we propose \method, which utilizes error-injected negative samples that have been largely overlooked in prior studies. 
Specifically, we define three strategic errors commonly occurring in LLMs and introduce a straightforward error-injected negative sample generation method to produce these target error types.
After that, \method employs an error-focused preference optimization objective, aiming to reduce the probability of negative samples and, in turn, minimize the likelihood of the introduced strategic errors, thereby enhancing overall model performance.
\method reduces the dependence on powerful models, requiring only minimal error injection ability and minimizes computational overhead by eliminating the scoring process, as shown in Figure \ref{intro}. 



Through extensive experimental validation, we demonstrate that \method effectively generates more erroneous samples compared to the original negative samples, thereby enhancing both overall and specific model capabilities, such as mathematical reasoning, code solving, and truthfulness, through negative sample augmentation with these strategic errors.
The results show improvements of up to 18.8\% in Llama3-8B-Instruct and 12.3\% in Qwen2.5-7B-Instruct on truthfulness, surpassing existing methods such as \citet{Cui2024UltraFeedbackBL}.
Additionally, ablation experiments highlight the significance of both error definition and error injection.

Our main contributions are as follows:
\begin{itemize}
    \item We introduce \method, which innovatively utilizes self-rewritten negative samples with preference optimiztation, effectively reducing the occurrence of undesirable error modes (\S \ref{sec:method}).

\item \method demonstrates exceptional versatility by achieving strong performance across a wide range of datasets and model scales (1.5B to 14B), proving its robustness and scalability in diverse settings (\S \ref{main_experiment}).

\item Our exploration of error definitions and the error injection reveals the impact of negative sample diversity, quality, and the severity of their errors on the final performance (\S \ref{sec:analysis}).

\end{itemize}

\section{Related Work}
%
\subsection{Preference Learning Data}
Many studies~\cite{guo2024directlanguagemodelalignment,li2024confidencemattersrevisitingintrinsic,chen2024selfplayfinetuningconvertsweak,kim2024aligninglargelanguagemodels} have recognized the value of self-generated data for model enhancement, typically using iterative improvement~\cite{liang2024isheepselfalignmentllmscratch,chen2024selfplayfinetuningconvertsweak,dong2024selfboostinglargelanguagemodels} or self-rewriting~\cite{li2024confidencemattersrevisitingintrinsic} to refine training sample quality and boost model performance.  UltraFeedback~\cite{Cui2024UltraFeedbackBL} constructs preference datasets via multi-model rejection sampling, selecting the highest-scoring response as positive and randomly rejecting others as negative. While existing methods focus on generating more correct answers, they struggle to improve the quality of erroneous samples~\cite{huang2024largelanguagemodelsselfcorrect,li2024confidencemattersrevisitingintrinsic}. 
Notably, \citet{huang-etal-2023-learning-preference} leverages the HHH Criteria to construct credibility-focused incorrect answers, but this approach is limited to TruthfulQA and not validated for reasoning or mathematical tasks.
Contemporaneous work~\cite{xu2024subtleerrorsmatterpreference} randomly introduced errors to enhance the results of a single task like math. 
\method introduces specific types of errors, improving the performance of tasks prone to such errors.

\subsection{Preference Learning Algorithm}
Preference learning trains models to align outputs with human preferences or goals, often using techniques like reward shaping or reinforcement learning to optimize performance and minimize unexpected behaviors. 
\citet{dpo} propose Direct Preference Optimization (DPO) to efficiently train large models for knowledge alignment using preference rankings instead of reward models. 
DPO optimizes classification loss from preference data, making implementing it simpler than reinforcement learning from human feedback. 
Recent improvements include: Smaug~\cite{pal2024smaug} uses the addition of regularity to improve the effectiveness of DPO. ORPO~\cite{Hong2024ORPOMP} strongly rewards the choice of answers by modifying the optimisation objective as well as removing the reference model. 
SimPO~\cite{meng2024simpo} reconstructs the training objective by adopting the length regularity reward.
KTO~\cite{kto} utilizes prospect theory to optimize model alignment with human feedback by directly maximizing a utility function based on binary signals, minimizing the need for preference data.
These algorithms sequentially increase the probability of positive samples and decrease the probability of negative samples. We focus on the data, leveraging it to optimize the final outcome by applying the same optimization method across varying datasets.

\begin{table*}[t]
\footnotesize
\centering
\renewcommand{\arraystretch}{1.8}
\scalebox{0.9}{
    \begin{tabularx}{1.1\textwidth}{l X X}
        \toprule
        \textbf{Strategic Errors} & \textbf{Descriptions} & \textbf{Tasks where This Error Commonly Occurs} \\ 
        \midrule
        
        Correctness  
        & Mistakes related to factual accuracy or calculations, such as incorrect facts, reasoning errors, translation issues, or improper use of tools and formulas.
        & Mathematical Calculation, Fact Verification, Text Summarization \\
        
        Logic  
        & Issues where the reasoning or argumentation is flawed, such as contradictions, unsupported conclusions, circular reasoning, or failure to follow a logical sequence in problem-solving or explanation.
        & Mathematical Reasoning, Commonsense Reasoning, Algorithm Design \\

        Hallucination  
        & Instances where the model generates information that is completely fabricated or false, without basis in reality or relevant data, often resulting in the creation of nonexistent facts or misleading details.
        & Knowledge-based QA, Code Generation, Machine Translation \\

        \bottomrule 
    \end{tabularx}
}
\caption{Strategic errors and their corresponding task types.}
\label{error_type}
\end{table*}

\section{Strategic Error Amplification}\label{sec:method}

\subsection{Overview}
\begin{figure*}[t]
    \centering
    \includegraphics[width=0.9\linewidth]{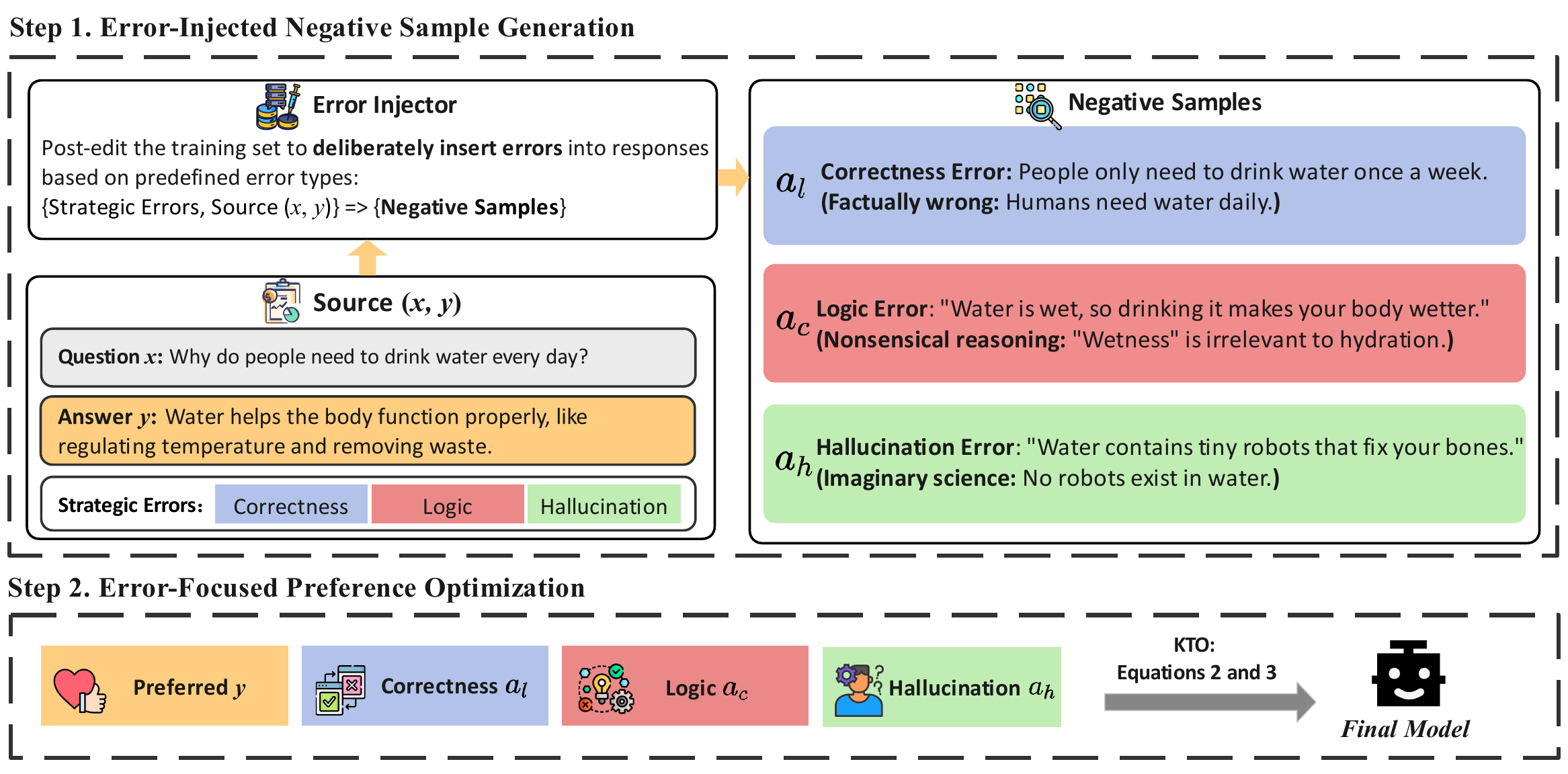}
    \caption{An overview of the \method. The entire process consists of two parts: 1) Error-Injected Negative Sample Generation. 2) Error-Focused Preference Optimization.}
    \label{overview}
\end{figure*}

In Figure \ref{overview}, \method consists of two components: Error-Injected Negative Sample Generation and Error-Focused Preference Optimization. 1) Error Generation: We predefine common LLM error types (correctness, logic, hallucination, termed Strategic Errors) and use the target model itself to post-edit training responses, intentionally injecting errors in specified proportions to address targeted capacity gaps. 2) Preference Optimization: Using KTO, we optimize both original positive samples and error-injected negative samples to refine the model’s ability to distinguish correct/incorrect responses, enhancing accurate answer generation.

%
    



\subsection{Definition of Strategic Errors}\label{subsec:strategic_errors}
Model development often involves analyzing problematic cases in specific scenarios to identify the types of issues that need to be addressed.
It is assumed that this process has been analyzed manually. 
We define strategic errors as inaccuracies, logical flaws, or misleading outputs that frequently occur in the responses generated by LLMs.
These errors are systematically identified based on common mistakes frequently observed in model outputs, enabling a more nuanced evaluation of their performance and robustness. 
Specifically, we categorize these strategic errors into three primary types: correctness errors, logic errors, and hallucinations. Table \ref{error_type} provides a detailed classification of these error types and highlights the tasks most susceptible to each category, offering a comprehensive overview of strategic errors and their relevance across different tasks.
By systematically identifying and analyzing these strategic errors, we aim to deepen our understanding of the limitations and capabilities of LLMs across various task domains. 

\subsection{Error-Injected Negative Sample Generation}

The goal of \method is to suppress the model's tendency to commit specific types of errors when answering questions in task-specific scenarios. To achieve this, constructing error-injected samples that closely resemble real-world mistakes becomes crucial.  
Negative samples that exhibit a moderate level of error, neither excessively incorrect nor overly accurate, can enhance the final outcomes of preference training.
We generate negative samples by leveraging the training set and predefined error types. Specifically, we prompt the LLM to modify the original correct answers by injecting errors corresponding to each identified error category, with the prompt we used detailed in Appendix \ref{sec:prompts}. This process can be formalized as follows:
Let $\mathcal{E}=\{e_c,e_l,e_h\}$ denote the set of predefined error types. For a given question $x$ and its correct answer $y$, we define a specific error instance $e\in\mathcal{E}$. The error-injected answer (negative sample) is then generated using the following function:
\begin{equation}
    a_{\text{error}} = \text{Injector}(x, y, e),
\end{equation}
where $a_{\text{error}}$ represents the answer with the injected error, and $e$ specifies the concrete error instance to be introduced into the question $x$. The error injection process is performed by the model itself, guided by the injector prompt. We utilize three distinct error types alongside the original question-and-answer pair to systematically inject errors. In this setup, the model generates responses tailored to each injected error type for every question type.

While our method is straightforward, the specific error categories are rooted in human empirical analysis. Currently, the construction of our prompts is generated by GPT-4o (see the specific prompt for generating the prompting words in Appendix \ref{sec:prompts}).
We also provide examples of the error-injected samples for each of the three error types in Appendix \ref{sec:cases}, which demonstrate the variations in the injected errors across different question types.
These three types of responses address the original question from different perspectives, yet each intentionally incorporates the respective error type. 
This process generates diverse negative samples simulating realistic errors, enhancing model robustness in preference optimization.

\subsection{Avoiding Pitfalls: Error-Focused Preference Optimization}

KTO~\cite{kto} is an effective algorithm that reduces the impact of negative samples in real-world scenarios. 
Although acquiring prompt-chosen-reject paired data remains challenging, it is still possible to evaluate output acceptability. 
Importantly, KTO's design eliminates the requirement for matched data pairs during training, making it particularly well-suited for scenarios focused on error cases.
At its core, KTO is grounded in the value function, which models human decision-making under uncertainty.
The value function captures the phenomenon of loss aversion, where humans are more sensitive to losses than equivalent gains.
The value of an outcome $z$ is calculated relative to a reference point $z_0$, with the function exhibiting concavity for gains and convexity for losses:
\begin{equation}
v(z; \lambda, \alpha, z_0) =
\begin{cases}
(z - z_0)^\alpha & \text{if } z \geq z_0 \\
-\lambda (z_0 - z)^\alpha & \text{if } z < z_0
\end{cases},
\end{equation}
where $\alpha$ controls risk attitudes, and $\lambda$ represents loss aversion.
The core objective of KTO is to optimize the expected utility of model generations:
\begin{equation}
    L_{KTO}(\pi_\theta, \pi_{\text{ref}}) = \mathbb{E}{x, y \sim D}[\lambda_y - v(r_\theta(x, y))],
\end{equation}
where $ r_\theta(x, y) = \log \frac{\pi_\theta(y|x)}{\pi_{\text{ref}}(y|x)} $ is the reward for output $y$ given input $x$, and $ v(r_\theta(x, y)) $ is the corresponding value based on the reward.
The reference model $\pi_{\text{ref}}$ serves as a baseline distribution (instruct model), guiding the model’s outputs while maintaining their alignment with human expectations.

\subsection{Complexity Discussion}

Typical negative sample construction uses rejection sampling~\cite{yuan2023scalingrelationshiplearningmathematical}, generating multiple responses per problem without explicit output pattern control, risking similar-quality outputs. For example, UltraFeedback~\cite{Cui2024UltraFeedbackBL} randomly constructs negatives via multi-model inference (Alpaca-GPT-4~\cite{alpaca-gpt4}, Llama~\cite{llama3}, Vicuna~\cite{vicuna}, $\geq$7B params), selecting three negatives and one positive per problem with a GPT4 scoring model, then randomly pairing one negative-positive pair for training. In contrast, our method eliminates the need for an auxiliary scoring model, drastically reducing data construction overhead. This optimization enhances both construction and generation efficiency.

\begin{table*}[t]
  \centering
  \scalebox{0.8}{
    \begin{tabular}{llcccccccc}
    \toprule
    \multirow{2}{*}{\textbf{Model}} & \multirow{2}{*}{\textbf{Method}} & \multicolumn{2}{c}{\textbf{Math}} & {\textbf{ Reasoning}} & \multicolumn{1}{c}{\textbf{Coding}} & {\textbf{Knowledge}} & \multicolumn{2}{c}{\textbf{Truthful}} & \multirow{2}{*}{\textbf{Avg.}} \\
    \cmidrule(lr){3-4} \cmidrule(lr){5-5} 
\cmidrule(lr){6-6} \cmidrule(lr){7-7} \cmidrule(lr){8-9} 
                      & & \textbf{MATH} & \textbf{GSM} & \textbf{BBH} &\textbf{HumanEval} &\textbf{MMLU} & \textbf{MC1} &\textbf{MC2} \\
    \midrule
        \multirow{3}{*}{Llama3-8B} &\multicolumn{1}{l}{Instruct} & 28.0 & 75.1 & \textbf{68.5} & 75.5 & \textbf{65.7} & 36.2 & 51.8 & 57.3 \\
    & \quad+Vanilla & \textbf{28.1} & \textbf{75.3} & 67.6 & 77.5 & 65.3 & 38.9 & 56.3 & 58.4 \\
    & \quad+\method & 25.8 & 72.2 & 62.9 & \textbf{79.5} & 64.7 & \textbf{55.0} & \textbf{66.1} & \textbf{60.9} \\
        \midrule
    \multirow{3}{*}{Qwen2.5-1.5B} & Instruct & 30.4 & 61.0 & \textbf{45.6} & \textbf{70.5} & \textbf{59.5} & 30.5 & 46.0 & 49.1 \\
    & \quad+Vanilla & 27.8 & 56.4 & 43.4 & \textbf{70.5} & 59.2 & 31.5 & 48.0 & 48.1
    \\
    & \quad+\method & \textbf{30.9} & \textbf{64.1} & 44.7 & 69.2 & 59.3 & \textbf{39.3} & \textbf{55.0} & \textbf{51.8} \\
    \midrule
    \multirow{3}{*}{Qwen2.5-7B} & Instruct & 47.2 & 72.5 & \textbf{67.5} & 88.7 & 72.8 & 42.4 & 59.3 & 64.3 \\
    & \quad+Vanilla & 46.8 & 73.4 & 59.4 & \textbf{90.9} & \textbf{72.9} & 51.0 & 65.3 & 65.7 \\
    & \quad+\method & \textbf{56.5} & \textbf{75.4} & 67.0 & 88.6 & 72.5 & \textbf{60.2} & \textbf{70.9} & \textbf{70.2} \\
 \midrule
    \multirow{3}{*}{Qwen2.5-14B} & Instruct & 50.1 & 85.6 & 73.7 & 89.1 & \textbf{79.7} & 51.5 & 69.2 & 71.3 \\
    & \quad+Vanilla & 50.1 & 86.8 & 73.1 & \textbf{89.2} & \textbf{79.7} & 52.0 & \textbf{69.6} & 71.5 \\
    & \quad+\method & \textbf{53.9} & \textbf{90.7} & \textbf{78.9} & 88.5 & 79.3 & \textbf{52.5} & 69.1 & \textbf{73.3} \\
    \bottomrule
    \end{tabular}}
    \caption{Main results on multiple test sets for multiple models. 
    The results show that our \method can deliver superior results compared to the existing method over multiple model series.}
  \label{tab:main_res}
\end{table*}

\section{Experiments}\label{main_experiment}
\subsection{Setup}
\paragraph{Baseline and Models}
Following previous work~\cite{meng2024simpo,dpo} of preference optimization, we use UltraFeedback~\cite{Cui2024UltraFeedbackBL} including 64K prompts and 256K responses (a prompt generates four responses, one of which is used as a positive sample and one of the remaining three is randomly selected as a negative sample), as the initial dataset for preference learning with KTO as a baseline.
We use the Qwen2.5 series (1.5B, 7B and 14B) and Llama3-8B-Instruct as the instruction models for our error-injecting before preference learning. We use the Qwen2.5-7B-Instruct as the base model for the analysis experiments.

\paragraph{Evaluations}
Math, reasoning, coding, knowledge, and truthfulness are core LLM capabilities, evaluated via out-domain benchmarks.
In the evaluations, we examine the model's performance in these areas, employing out-domain benchmarks to assess its ability to assimilate math (MATH~\cite{hendrycks2021measuringmathematicalproblemsolving} and GSM~\cite{gsm}), executing complex reasoning (BBH~\cite{bbh}), coding (HumanEval~\cite{humaneval}), knowledge (MMLU~\cite{MMLU}) and truthful~\cite{lin-etal-2022-truthfulqa} tasks. 
We use the exact match as the evaluation metric to assess the models’ ability of {Math} and {Reasoning} to provide correct answers.
We use the HumanEval to evaluate the models’ capability to generate functionally correct programs from docstrings, with pass@10 as the evaluation metric.
We employ the MMLU to measure factual {knowledge}, using accuracy as the evaluation metric.
For {Truthful}, since there are multiple correct answers, apart from standard accuracy (MC1), we also report the ratio of likelihood sum of correct answers overall candidate answers (MC2).

\paragraph{Training Details}
The model was trained using a learning rate of $5 \times 10^{-6}$, following a cosine decay strategy for the learning rate schedule.
A per-device batch size of 2 was used during training, and to achieve an effective total batch size of 64, we employed gradient accumulation over 8 steps, which optimized memory usage during training.
The training process utilized a multi-GPU distributed setup with 4 GPUs and was initialized with a random seed of 42 for reproducibility.
For optimization, we used the Adam optimizer~\cite{kingma2014adam} with $\beta_1 = 0.9$, $\beta_2 = 0.999$, and $\epsilon = 1 \times 10^{-8}$.
Training proceeded for a total of 5731 steps, determined by the total number of training samples and the batch size.
Regarding the hyperparameters of KTO, we set $\beta=0.1$.
Additionally, we fine-tuned the model in all experiments using LoRA~\cite{hu2022lora}.
The LoRA configuration included a rank of 8, and a scaling factor $\alpha$ of 16, without dropout.
\subsection{Main Results}

Table \ref{tab:main_res} demonstrates that constructing diverse negative samples for preference optimization is effective, as reducing the occurrence of error patterns leads to improved model performance. Compared to the original instruct model and UltraFeedback's KTO training, our \method consistently outperforms across different models, with notable improvements on error-rich datasets such as MATH and TruthfulQA. For instance, Qwen2.5-7B achieves gains of +10 points on MATH and +12 points on TruthfulQA. These improvements are consistent across various architectures (e.g., Llama3-8B, with a +20 point increase) and model scales. Smaller models ($<$8B) benefit the most from error-type injection to reduce hallucinations, while larger models ($\geq$8B) show enhanced reasoning capabilities through the reduction of incorrect patterns. 

Notable declines in some models (e.g., Qwen2.5-1.5B 3 points in Math) may stem from low-quality injected errors or limited capabilities, as further analyzed in Tables \ref{source_rewriter} and \ref{sample_quality}.
Minimal changes occur in strong areas like coding (scores 90) and knowledge-based MMLU. 
Knowledge-related capabilities are primarily determined during the pre-training phase, with post-training having only a limited impact, typically resulting in changes of less than two points~\cite{wang2023far,yang2024self}. 
For code-related tasks, baseline performance is already high, and the evaluation set contains only 164 samples, making the scores highly sensitive to individual errors. Consequently, variations within two points are better interpreted as random fluctuations rather than meaningful changes. In contrast, other tasks, such as GSM, include larger evaluation sets (e.g., 1,319 samples), where noticeable score improvements (e.g., +10 points) require consistently correct responses across many instances. These differences in dataset size and difficulty explain the observed score variability in code and knowledge.

\begin{table*}[t]
    \centering
    \scalebox{0.9} {
    \begin{tabular}{llcccccccc}
        \toprule
        \multirow{2}{*}{\textbf{ID}} & 
        \multirow{2}{*}{\textbf{Error Type}} & 
        \multicolumn{2}{c}{\textbf{Math}} & 
        \multicolumn{1}{c}{\textbf{Reasoning}} & 
        \multicolumn{1}{c}{\textbf{Coding}} & 
        \multicolumn{1}{c}{\textbf{Knowledge}} & 
        \multicolumn{2}{c}{\textbf{Truthful}} & 
        \multirow{2}{*}{\textbf{Avg.}} \\
        
        \cmidrule(lr){3-4} \cmidrule(lr){5-5} \cmidrule(lr){6-6} \cmidrule(lr){7-7} \cmidrule(lr){8-9} 
        & & \textbf{MATH} & \textbf{GSM} & \textbf{BBH} & \textbf{HumanEval} & \textbf{MMLU} & \textbf{MC1} & \textbf{MC2} \\
        
        \midrule
        1 & None (Instruct)   & 47.2 & 72.5 & 67.5 & {88.7} & 72.8 & 42.4 & 59.3 & 64.3 \\
        2 & Untargeted        & 56.6 & 70.8 & 68.6 & 87.6 & \underline{72.9} & 55.6 & 67.4 & 68.5 \\ 
        
        \midrule
        3 & Logic             & 54.5 & 71.8 & 65.2 & 89.0 & 72.7 & 59.5 & 70.7 & 69.1 \\
        4 & Correctness       & \underline{54.7} & 69.6 & 65.4 & \underline{89.3} & 72.8 & 57.8 & 68.8 & 68.3 \\
        5 & Hallucination     & 53.7 & 74.4 & 66.7 & \underline{89.3} & 72.5 & 60.8 & \underline{72.0} & 69.9 \\
        
        6 & 3+4 & 52.9 & \textbf{79.8} & \underline{69.4} & 87.7 & \textbf{73.2} & 52.6 & 64.9 & 68.6 \\
        7 & 3+5 & 52.5 & {77.5} & 67.4 & \underline{89.3} & 72.6 & \textbf{62.8} & \textbf{72.9} & \textbf{70.7} \\
        8 & 4+5 & 53.9 & 75.7 & {67.7} & 88.0 & 72.7 & \underline{61.8} & 71.7 & \underline{70.2} \\
        9& {Multi}& 55.8&\underline{79.5}&\textbf{70.4}&\textbf{89.6}&\textbf{73.4}&50.2&64.7&69.1\\
        10 & 3+4+5 & \textbf{56.5} & 75.4 & 67.0 & 88.6 & 72.5 & 60.2 & 70.9 & \underline{70.2} \\
        
        
        \bottomrule
    \end{tabular}}
    
    \caption{Results for different error types on each tasks. Training on the most frequent error type improves task performance. 
    Math, code, and reasoning tasks, are prone to correctness and logical errors, while TruthfulQA, a knowledge-based task, is more susceptible to hallucinations. Constructing the relevant errors enhances performance in each task. The best results are bolded and the second best results are underlined.}
    \label{tab:erro_type_results}
\end{table*}

\section{Analysis}\label{sec:analysis}
\subsection{Analysis on Strategic Error Definition}
\paragraph{Setup}
Table \ref{tab:erro_type_results} presents a comparison of the effects of different error types across multiple tasks. The evaluation includes baselines: ID 1 (None), representing the original instruction-tuned model, and ID 2 (Untargeted), which corresponds to results obtained by rewriting positive samples without type-specific constraints. 
We report results for three error types (IDs 3–5) and their mixed variants (IDs 6–10).
IDs 6–8 and ID 10: These involve mixing responses from different question types after injection in equal proportions, with the ratio order specified as logic, correctness, and hallucination. For example: ID 7 = 50\% logic + 50\% hallucination (a 2-type mix).
ID 9: A distinct mixed variant where responses from all three error types (logic, correctness, hallucination) are combined to generate a single integrated type.
\paragraph{Effect of Different Error Types}
The results show that even without introducing specific error types, the model can still improve, but the improvement is mainly concentrated in three dimensions: MATH, BBH, and Truthfulness. 
When errors are introduced in the logic or correctness type, the performance in MATH, code-related tasks, and truthfulness will be enhanced. 
When introducing hallucination errors, the improvement in credibility is more significant compared to the previous two error types. 
We observe that the introduction of hallucinations is often associated with noticeably shorter model outputs, as illustrated in Appendix~\ref{sec:cases}. 
Reasoning tasks typically benefit from a step-by-step thought chain approach. 
Such short responses can adversely affect performance on complex reasoning tasks, where extended generation length is often necessary for success. 
Prior studies ~\cite{li202512surveyreasoning,guo2025deepseek} have emphasized that longer reasoning chains are positively correlated with improved performance on more challenging reasoning problems.
These results demonstrate the necessity of introducing problem types and the influence of problem-type diversity on the comprehensive performance of multiple tasks. 
We present additional evaluation examples of error handling for the three types of baseline errors in Appendix \ref{sec:cases}, demonstrating the effectiveness of \method.

\begin{table}[t!]
    \centering
    \scalebox{0.8} {
        \begin{tabular}{lrrrr}
            \toprule
            \multirow{2}{*}{\textbf{Error Type}} 
            & \multicolumn{2}{c}{\textbf{Qwen2.5-72B}} & \multicolumn{2}{c}{\textbf{Llama3.1-70B}} \\
            \cmidrule(lr){2-3} \cmidrule(lr){4-5}
            & \textbf{Source} & \textbf{Injected} & \textbf{Source} & \textbf{Injected} \\
            \midrule
            Logic & 11,249 & \textbf{45,815} & 9,891 & \textbf{43,044} \\
            Correctness & 12,367 & \textbf{43,577} & 10,307 & \textbf{38,894} \\
            Hallucination & 6,515 & \textbf{47,069} & 8,582 & \textbf{46,995} \\
            \bottomrule
        \end{tabular}
    }
    \caption{Comparison of error types in the original answers and those after error injection. The number of specific error types in the modified answers increased compared to the original answers.}
    \label{tab:prelim}
\end{table}

\paragraph{Effect of Mixing Strategy}
In Table \ref{tab:erro_type_results}, we conducted a preliminary exploration of different error type ratios of mixing. 
The underlying philosophical distinctions are reflected in the introduction of different error types and their varying impacts across tasks. All hallucination-related error types lead to substantial performance improvements on TruthfulQA, a benchmark specifically designed to evaluate sensitivity to hallucination.

Mixing error types improves overall performance: most average results rank top-2, and more mixed types enhance nearly all subtasks (though less than single-task gains, e.g., GSM/BBH for ID 6/10). 
Results show that multi-error injection (ID 9) boosts scores (+4.8 avg, +9.0 MATH, +7.0 GSM), confirming effectiveness, though ID 10 outperforms.
For generalization, we use ID 10 (full mixing).
The introduction of specific errors led to notable improvements in the corresponding evaluation results. For instance, errors involving hallucination types, such as those in IDs 5, 7, 8, and 10, resulted in significant gains compared to cases without explicit hallucination injection (e.g., IDs 1 and 6), with the largest improvement nearing 10 points.
When focusing on specific capabilities, such as mathematics and reasoning, a blend of logic and correctness is most effective. For truthfulness, combining logic with hallucination errors yields better results. In general, mixing a broader range of error types can lead to more comprehensive performance improvements across various tasks.


\subsection{Analysis on Error-Injected Negative Sample Generation}
\paragraph{Effect of Injected Error Quality}
To assess whether the model's error injection mechanism can generate responses with specific error types based on the original response, we conducted evaluations using models such as Llama3.1-70B and Qwen2.5-72B. The results are presented in Table \ref{tab:prelim}.
We employed these evaluation models to classify responses based on the question, reference answer, and rejection, performing binary classification to determine if the response exhibits a specific error type. 
The evaluation prompt is provided in Appendix \ref{sec:prompts}.
The negative samples in the original dataset may not consistently exhibit greater error severity than their corresponding positive samples. In Table~\ref{tab:prelim}, our generated negative samples demonstrate a higher degree of incorrectness compared to the original set. 
The presence of less erroneous examples in the original negative samples likely hinders the effectiveness of preference training, potentially degrading overall model performance.

\begin{table}[t!]
\centering
\scalebox{0.8} {
\begin{tabular}{lcccc}
\toprule
& \multicolumn{2}{c}{\textbf{Qwen2.5-72B}} & \multicolumn{2}{c}{\textbf{DeepSeek-V3}} \\
\cmidrule(lr){2-3} \cmidrule(lr){4-5}
& \textbf{Score} & \textbf{Win-Rate} & \textbf{Score} & \textbf{Win-Rate} \\\midrule
UltraFeedback & 6.87 & - & 7.09 & - \\
Correctness & 6.91 & 0.68 & 7.13 & 0.69 \\
Logic & 6.67 & 0.67 & 6.98 & 0.67 \\
Hallucination & 6.65 & 0.67 & 7.00 & 0.62 \\
Minor & 7.30 & 0.71 & 7.43 & 0.73 \\
Moderate & 7.05 & 0.70 & 7.22 & 0.67 \\
Major & 6.51 & 0.64 & 6.93 & 0.60 \\
Critical & 3.46 & 0.30 & 4.73 & 0.25 \\
\bottomrule
\end{tabular}
}
\caption{Comparison of negative sample quality. The negative samples with error injection show no significant quality degradation compared to the original ones.}
  \label{tab:gen_data}
\end{table}
Table~\ref{tab:gen_data} presents a comparison of error correction performance across four severity levels (minor, moderate, major, critical) and three error categories (factual correctness, logical consistency, and hallucination). The evaluation is conducted on 1,000 sampled instances using two scoring methodologies: direct absolute scoring~\cite{kim2024prometheus} and relative win rate~\cite{vicuna}, applied to Qwen2.5-72B-Instruct and DeepSeek V3.
Notably, the performance gap becomes substantial only when the correction quality is critically poor, with direct scores dropping from 6.87 to 3.46. Despite this, our proposed error-specific negative samples maintain performance comparable to original baselines across both evaluation methods and models, showing a direct score difference of no more than 1 point and a win rate difference of no more than 0.1. These results demonstrate that our negative samples are effective for model training without compromising output quality.
 \begin{table}[t!]
\centering
\scalebox{0.8}{
\begin{tabular}{lrrcr}
\toprule
    \multirow{2}{*}{\textbf{Performance}} & 
    \multirow{2}{*}{\textbf{Baseline}} & \multicolumn{3}{c}{\textbf{Error Injector}} \\
    \cmidrule(lr){3-5} 
    & & \textbf{1.5B} & \textbf{7B (Self)} & \textbf{72B} \\ 
    \cmidrule(lr){1-5}
    MATH  & 47.2 & 55.6 & \textbf{56.5} & 55.0 \\
    GSM   & 72.5 & 77.0 & 75.4 & \textbf{83.5} \\
    BBH   & 67.5 & \textbf{67.6} & 67.0 & 66.1 \\
    HumanEval & \textbf{88.7} & 86.9 & 88.6 & 88.1 \\
    MMLU  & 72.8 & \textbf{73.1} & 72.7 & 72.4 \\
    Truthful-MC1 & 42.4 & 44.8 & \textbf{60.2} & \textbf{60.2} \\
    Truthful-MC2 & 59.3 & 59.8 & 70.9 & \textbf{71.7} \\
    \midrule
    Avg.  & 64.3 & 66.4 & 70.2 & \textbf{71.0} \\
\bottomrule
\end{tabular}
}
\caption{Effect of Source of the Generated Error. Larger models have stronger error-injection capabilities. }
\label{source_rewriter}
\end{table}

\begin{table}[t!]
\centering
\scalebox{0.7}
{
\begin{tabular}{lccccc}
\toprule
    \multirow{1}{*}{\textbf{Task}} &\centering\textbf{Baseline}&\centering\textbf{Minor} & \centering\textbf{Moderate}&{\centering\textbf{Major}}&{\centering\textbf{Critical}}\\ \midrule
    MATH & 47.2 & \textbf{57.3} & 56.7 & 56.6 & 53.3 \\
    GSM & \textbf{72.5} & 64.4 & 67.6 & 70.8 & 62.1 \\
    BBH & 67.5 & 67.7 & 68.3 & 68.6 & \textbf{69.6} \\
    HumanEval & \textbf{88.7} & 88.2 & 87.7 & 87.6 & 86.0 \\
    MMLU & 72.8 & \textbf{73.2} & \textbf{73.2} & 72.9 & 73.0 \\
    Truthful-MC1 & 42.4 & 49.8 & 52.3 & \textbf{55.6} & 45.7 \\
    Truthful-MC2 & 59.3 & 64.0 & 66.1 & \textbf{67.4} & 61.3 \\
    \midrule
    Avg. & 64.3 & 66.4 & 67.4 & \textbf{68.5} & 64.4 \\
\bottomrule
\end{tabular}
}
\caption{Comparison of the effects obtained by negative samples with different error levels. Models with the error level of ``Major'' work best, indicating that negative samples require recognizable errors.}\label{sample_quality}
\end{table}

\paragraph{Effect of Source of the Generated Error}
We show in Table \ref{source_rewriter} the effect of different sizes (1.5B, 7B, and 72B) of error-injected negative samples of the same model architecture on the final training effect of the model, which shows that the models of different sizes have different abilities to introduce errors by prompt. 
This ability to introduce errors is relatively poor for small models and strong for large models.
However, we find that even with small models, after error injection, the model's can all be further improved, and the gap between these scores is not very large, which indicates that \method is not very demanding for the ability to correct errors, and further illustrates the robustness.

\paragraph{Effect of Severity of Errors}
Table \ref{sample_quality} analyzes the impact of error severity on training, classifying introduced errors into four levels: Minor (few errors), Moderate (medium errors), Major (many errors), and Critical (almost entirely incorrect). 
Results show that prompting models to introduce errors generally improves performance via preference learning, except for Critical-level errors, where improvement is negligible. This highlights that negative samples must maintain a reasonable quality for the model to distinguish errors effectively. 
\method achieves improvements by reducing the likelihood of negative samples: our negative samples contain more errors than original responses while maintaining comparable quality.{~\citet{xu2024subtleerrorsmatterpreference} can be categorized as introducing Major errors. Nevertheless, the performance resulting from these randomly injected errors remains inferior to that achieved through the introduction of targeted error types, as implemented in \method.}

\begin{table}[t!]
    \centering
    \scalebox{0.8}{
        \begin{tabular}{lcccc}
            \toprule
            \multirow{2}{*}{\textbf{Instruct}} & \multicolumn{2}{c}{\textbf{RoN}} & \multicolumn{2}{c}{\textbf{SeaPO}} \\
            \cmidrule(lr){2-3} \cmidrule(lr){4-5} & \textbf{DPO} & \textbf{KTO} & \textbf{DPO} & \textbf{KTO} \\
            \midrule
            64.3 & 64.8 & 65.7 & 69.4 & \textbf{70.2} \\
            \bottomrule
        \end{tabular}
    }
    \caption{Impact of optimization objectives on average performance. ``RoN'' random selects one from three negative samples as the rejection (original UltraFeedback), while \method selects from the rewritten negatives.}\label{tab:obj}
\end{table}

\subsection{Analysis on Error-Injected Preference Optimization}
In Table \ref{tab:obj}, we present the average results across the five dimensions for both the data we constructed and the original data, evaluated under two optimization objectives. 
DPO is more susceptible to the influence of noisy data~\cite{dpo}, where both positive and negative samples may contain correct responses, a problem that is particularly evident in the RoN (random selection of one of the remaining three samples using the original dataset). 
After training on the original data, the model performance showed slight volatility, while KTO demonstrated better generalization, with results showing a modest improvement. 
However, utilizing our generated, containing more injected errors, we observed an improvement in the DPO results (almost +4 points). 
When combined with KTO, which is less sensitive to sample quality, even better performance was achieved. 
This demonstrates the robustness of \method
in different learning algorithms.

\section{Conclusion}
Due to model limitations and no explicit quality control in preference data construction, positive and negative samples may have similar quality; in some cases, negative samples may even be better.
To address this issue, we propose \method that explicitly introduces strategically defined errors. 
Applying preference optimization reduces error probabilities and enhances overall performance.
Evaluations across five capability dimensions and multiple models (1.5B-14B) demonstrate the robustness of our approach.
These strategic errors vary depending on the task, and targeting the most error-prone types can significantly boost task performance.

\section*{Limitations}
There are several limitations to our work.
First, our experiments are constrained by available resources, particularly GPU memory capacity, which limits the number of models we can test, especially at larger scales (e.g., 70B models). Additionally, we did not define a broader range of error types, such as those related to grammar or safety, which could potentially enhance the performance of specific tasks.
Another limitation is the lack of targeted error analysis for specific task types. As a result, the three error categories we defined may not fully capture less common error types that occur in certain tasks.
Finally, our experiments were conducted using only one preference dataset. The quality of positive samples in different datasets may introduce variations in the final model results.
\section*{Ethics Statement}
We take ethical considerations very seriously and strictly adhere to the ACL Ethics Policy.
All experiments are
conducted on open datasets and the findings and
conclusions of this paper are reported accurately
and objectively.

\section*{Acknowledgments}
This work was supported in part by Guangdong S\&T Program (Grant No. 2024B0101050003), Guangdong Basic and Applied Basic Research Foundation (Grant No. 2024A1515011491), and Shenzhen Science and Technology Program (Grant Nos. ZDSYS20230626091203008, KJZD20231023094700001, KQTD20240729102154066). 
We would like to thank the anonymous reviewers and meta-reviewer for their insightful suggestions.

\bibliography{custom}

\clearpage

\appendix

\section{Appendix}\label{sec:appendix}

\subsection{Prompts}\label{sec:prompts}
Multiple experiments, specifically, the results in Table~\ref{tab:erro_type_results} and the Moderate and Major error categories in Table~\ref{source_rewriter}, demonstrate that, despite the simplicity and diversity of the prompt construction (including both prompts with specific error types and those without), our method consistently yields significant performance gains. In several tasks across different model architectures, improvements of up to 5 percentage points are observed. 

We provide the prompt templates used in our study. 
The first template, titled \textit{Prompt Template for Error-Injected Negative Sample Generation}, is used for error injection. This template directs models to deliberately modify a given response in order to introduce a specified error type. It includes fields for the question, original response, error type, and error description. Models are then instructed to revise the response intentionally to incorporate the error, without providing additional explanation.
The second template, titled \textit{Prompt Template for Error Detection}, is used for error detection. This template directs evaluators to determine if a given answer contains the specified error. Evaluators are provided with the question, answer, error type, and error description, and are required to respond with ``yes'' or ``no'' based on the presence of the error, without further elaboration.
The third template, \textit{Prompt Template for Error Injection}, is used to generate responses containing injected errors. The fourth template, \textit{Prompt Template for Error Injection by Severity Level}, is designed to produce prompts corresponding to three levels of error severity: Minor (Level 1), Moderate (Level 2), and Major (Level 3), without specifying error types. For prompts related to evaluating the quality of generated responses, refer to the ``Prompt Scoring'' and ``Prompt Comparison sections''.

\subsection{Case Studies}\label{sec:cases}

\paragraph{Error Injection Cases}
The following examples illustrate error injection in three distinct cases.

For the correctness case, the statement ``Exams are the only way to measure a person’s knowledge'' is factually incorrect because it disregards valid alternative assessment methods, such as course grades, project outcomes, and portfolio reviews. The claim that exams are the sole tool for evaluation contradicts established practices in educational assessment, making the statement inaccurate. This error arises from providing information that does not align with factual knowledge.

For the logic case, the misinterpretation of Thoreau’s metaphor becomes apparent when the response states that someone cannot keep pace because they ``cannot hear the same drummer.'' The original metaphor emphasizes individuality (``hearing a different drummer''), but this interpretation shifts the meaning to one of auditory limitation. By altering the intended message, the response loses coherence and fails to preserve the logical structure of the argument. Such errors occur when the reasoning process distorts or misrepresents the relationships between ideas.

For the hallucination case, the fabrication of fictional concepts like ``tiny invisible fairies'' and ``fairy theory'' introduces imaginary elements with no basis in reality or historical context. Associating these fictitious ideas with Descartes’ era, despite the absence of any scientific or historical grounding, demonstrates how the model generates content that does not exist. This type of error results from inventing concepts or details entirely unsupported by evidence, leading to responses that include hallucinatory content.

\paragraph{Evaluation Cases}  
Based on the evaluation results, we present case studies to compare the baseline model (Qwen2.5-7B-Instruct) with our fine-tuned model across three error categories: correctness, logic, and hallucination. In each case, the baseline answer refers to the original model’s output, while our answer represents the improved model’s response. Incorrect components of the baseline responses are highlighted in red, while the corrected sections of our model’s responses are also marked in red. These cases illustrate the reduction in errors achieved by our fine-tuned model compared to the baseline.

For correctness errors, the baseline model incorrectly applied the order of operations, leading to an erroneous result. Our model rectified this by adhering to the correct sequence of calculations.

For logic errors, the baseline model erroneously calculated the interest for a single phone and propagated this mistake to the final result. Our model identified and addressed the reasoning flaw.

For hallucination errors, the baseline model misinterpreted the definition of the hypotenuse’s height, introducing incorrect information. Our model corrected this by applying domain-specific knowledge.


\onecolumn

\begin{tcolorbox}[colback=red!5!white, colframe=red!75!black, title=Prompt Template for Error-Injected Negative Sample Generation]
    Your task is to deliberately modify the provided response to introduce the specified error. \\ \\
    \#\#\# Task:\\
    Analyze the given question, original response, error type, and error description. Then, revise the response to intentionally include the specified error. \\ \\
    1. **Question**: \{question\} \\
    2. **Original Response**: \{response\} \\
    3. **Error Type**: \{error\_type\} \\
    4. **Error Description**: \{error\_description\} \\ \\
    \#\#\# Instructions: \\
    - Modify the original response to clearly incorporate the specified error.\\
    - Do not include any explanations, notes, or other text in your output.\\
    - Output only the revised response. \\ \\
    \#\#\# Revised response:
\end{tcolorbox}

\begin{tcolorbox}[colback=cyan!5!white, colframe=cyan!75!black, title=Prompt Template for Error Detection]
    You are an AI evaluator. I will give you a question, its answer, and a specific error type with a description. Your task is to determine if the answer contains the described error. Please respond with "yes" if the answer contains the error and "no" if it does not. Only respond with "yes" or "no" without any further explanation. \\ \\
    Question: "\{question\}" \\
    Answer: "\{answer\}" \\
    Error Type: "\{error\_type\}" \\
    Error Description: "\{error\_description\}" \\ \\
    Does the answer contain the described error?
\end{tcolorbox}

\begin{tcolorbox}[colback=black!5!white, colframe=black!75!black, title=Prompt Template for Generation of Error-Injection]
    Please help me design a prompt in Python string format. I will input the question, its correct answer, an error type, and a brief description of that error type. The model needs to intentionally modify the correct answer to incorporate the specified error type, outputting only the modified incorrect answer without any extra content. The Python string prompt should include placeholders for the input information.

\end{tcolorbox}

\begin{tcolorbox}[colback=green!5!white, colframe=green!75!black, title=Generation of Error-Injection Prompts of Different Levels]
    Please help me design a prompt in the form of a Python string. I will input a question, its correct answer, an error severity level, and a brief description of that severity level. The model needs to intentionally modify the correct answer into an incorrect one that matches the specified error severity. The output should only contain the modified incorrect answer, with no additional content. The Python string prompt must include placeholders for the input information.

\end{tcolorbox}

\begin{tcolorbox}[colback=black!5!white, colframe=black!75!black, title=Prompt Scoring]
   \#\# Task: 
Evaluate the quality of an answer while COMPLETELY IGNORING factual errors.

\#\# Note: 
The answer may contain inaccuracies or technical errors. DO NOT consider correctness in scoring.

\#\# Dimensions (1-10 scale):
1. Logical Structure - Organization and reasoning flow
2. Language Fluency - Grammar and readability
3. Topic Relevance - Alignment with question intent
4. Depth - Detail level and analysis complexity
5. Consistency - Self-contradiction check

\#\# Input:
Question: {question}
Answer: {answer}

\#\# Output Format:
Single integer between 1-10 calculated as: 
(Round average of all dimension scores to nearest integer)

ONLY OUTPUT THE FINAL INTEGER

\end{tcolorbox}

\begin{tcolorbox}[colback=black!5!white, colframe=black!75!black, title=Prompt Comparison]
   Compare two answers (A/B) while COMPLETELY IGNORING FACTUAL ERRORS. Determine which response demonstrates superior quality in non-factual dimensions.

\#\# Critical Rules:
1. DO NOT CONSIDER: Accuracy, technical correctness, or truthfulness
2. EVALUATE ONLY:
   Structural integrity of arguments
   Language mechanics and readability
   Relevance to question intent
   Depth of analysis
   Internal consistency
3. Tie only if indistinguishable across all dimensions

\#\# Assessment Protocol:
1. Read both answers fully
2. Identify potential errors → Explicitly exclude from evaluation
3. Compare dimension-by-dimension:
   a) Logical Structure: Which has clearer reasoning flow?
   b) Language Fluency: Which is more professionally written?
   c) Topic Relevance: Which stays focused on core question?
   d) Depth: Which provides more layered analysis?
   e) Consistency: Which avoids self-contradictions?
4. Declare winner with majority advantages

\#\# Input:
Question: \{question\}
Answer A: \{answer\_a\}
Answer B: \{answer\_b\}

\#\# Output:
Strictly ONE of these options:
A | B | tie

DO NOT OUTPUT ANY OTHER TEXT OR EXPLANATION.
\end{tcolorbox}

\begin{tcolorbox}[colback=yellow!5!white, colframe=yellow!75!black, title=Case of Correctness Error Injection]
    \textbf{Question:}
    
    Check if the following statement is true or false: 
    
    Exams are the best way to measure a person's knowledge
    \\\\
    \textbf{Correct answer:}

    False.
    Exams are not the only way to measure a person's knowledge.
    Other ways include course grades, project outcomes, problem-solving tasks, interview questions, portfolio reviews, and letters of recommendation.
    Exams are one tool used to measure knowledge, but they should not be the only tool relied on.
    \\\\
    \textbf{Answer injected with correctness error:}

    False.
    Exams are \textcolor{red}{the only way} to measure a person's knowledge.
    Other methods such as course grades, project outcomes, problem-solving tasks, interview questions, portfolio reviews, and letters of recommendation are \textcolor{red}{invalid}.
    Exams are \textcolor{red}{the definitive tool} for assessing knowledge.
 
\end{tcolorbox}

\begin{tcolorbox}[colback=pink!5!white, colframe=pink!75!black, title=Case of Logic Error Injection]
    \textbf{Question:}
    
    Thoreau’s metaphors are highly visual. Though they’re clever and original, they aren’t far-fetched. Thoreau takes his comparisons from nature and from other things with which he and his audience are familiar.
    To be sure you understand Thoreau’s figures of speech, paraphrase the following metaphor.
    Do his metaphors apply to readers today?
    
    ``If a man does not keep pace with his companions, perhaps it is because he hears a different drummer. Let him step to the music which he hears, however measured or far away.''
    \\\\
    \textbf{Correct answer:}

    In this metaphor, Thoreau is suggesting that if a person doesn't conform or fit in with others around them, it might be because they have their own unique path and aspirations. He encourages the individual to follow their own internal guidance and passions, no matter how different or distant it may seem from the norm.
    
    Thoreau's metaphors can still apply to readers today, as it highlights the importance of individuality and staying true to oneself in a world that often values conformity.
    \\\\
    \textbf{Answer injected with logic error:}

    In this metaphor, Thoreau is suggesting that if a person doesn't keep pace with others, it is because they are \textcolor{red}{deaf to the same drummer}. He encourages the individual to follow their own internal guidance and passions, no matter how different or distant it may seem from the norm.
 
\end{tcolorbox}

\begin{tcolorbox}[colback=green!5!white, colframe=green!75!black, title=Case of Hallucination Error Injection]
    \textbf{Question:}
    
    I am refering to earlier, back in Descartes age
    
    ``If a man does not keep pace with his companions, perhaps it is because he hears a different drummer. Let him step to the music which he hears, however measured or far away.''
    \\\\
    \textbf{Correct answer:}

    Back in Descartes' age, there was an idea that the physical world was composed of atoms that were arranged in a certain way and that this arrangement formed the foundation of all knowledge.
    This is now known as the atomic theory and it is one of the core principles of modern science.
    \\\\
    \textbf{Answer injected with hallucination error:}

    Back in Descartes' age, there was an idea that the physical world was composed of \textcolor{red}{tiny invisible fairies} that danced in a certain way and that this dance formed the foundation of all knowledge. This is now known as the \textcolor{red}{fairy theory} and it is one of the core principles of modern science.
 
\end{tcolorbox}

\begin{tcolorbox}[colback=yellow!5!white, colframe=yellow!75!black, title=Case of Correctness Errors in Evaluation]
    \textbf{Question:} 
    
Find $\frac{1}{-6} \cdot 6 \cdot 7 + 8 \cdot 2 \div 8 \cdot (7-1)$.
    \\\\
    \textbf{Baseline answer:} 

   First, we simplify the expression inside the parentheses: $7-1 = 6$.
Next, we perform the multiplication and division from left to right:
\[
\frac{1}{-6} \cdot 6 \cdot 7 = -1 \cdot 7 = -7,
\]
and
\[
8 \cdot 2 \div 8 = 16 \div 8 = 2.
\]
\textcolor{red}{Then, we multiply the results of the two parts:
\[
-7 \cdot 2 = -14.
\]
}
Finally, we add the result to the remaining part of the expression:
\[
-14 + 8 \cdot 2 = -14 + 16 = 2.
\]
Therefore, the final answer is $\boxed{2}$.
Final Answer: The final answer is $2$. I hope it is correct.
    \\\\
    \textbf{Our answer:} 

    First, we simplify the expression inside the parentheses: $7-1=6$. Then, we simplify the division and multiplication: $8 \cdot 2 \div 8 = 2$. So, the expression becomes

$$\frac{1}{-6} \cdot 6 \cdot 7 + 2 \cdot 6 = -7 + 12 = \boxed{5}.$$
Final Answer: The final answer is $5$. I hope it is correct.

\end{tcolorbox}

\begin{tcolorbox}[colback=pink!5!white, colframe=pink!75!black, title=Case of Logic Errors in Evaluation]
    \textbf{Question:} 

    Shiela bought five cell phones for \$150 each for a 3-month installment. A 2\% interest will be charged for each unit. How much will Shiela pay each month for 3 months?
    \\\\
    \textbf{Baseline answer:} 
    
     First, we need to calculate the total cost of the cell phones before interest. 5 cell phones at \$150 each is 5 x 150 = \$750. The interest for each cell phone is 2\%, \textcolor{red}{so for 5 cell phones, the total interest is 5 x 750 x 0.02 = \$75.} The total cost including interest is 750 + 75 = \$825. Since she is paying this over 3 months, we divide the total cost by 3: 825 / 3 = \$275. So the answer is 275. Each month, Shiela will pay \$275. 
    \\\\
    \textbf{Our answer:}
    
     The total cost of the cell phones is 5 x \$150 = \$750. The interest for each unit is 2\% of \$150, which is 0.02 x \$150 = \$3. \textcolor{red}{So the total interest for all the cell phones is 5 x \$3 = \$15.} The total amount to be paid is \$750 + \$15 = \$765. The monthly payment for 3 months is \$765 / 3 = \$255. So the answer is \$255.

\end{tcolorbox}

\begin{tcolorbox}[colback=green!5!white, colframe=green!75!black, title=Case of Hallucination Errors in Evaluation]
    \textbf{Question:} 
    
    In a certain isosceles right triangle, the altitude to the hypotenuse has length $4\sqrt{2}$.  What is the area of the triangle?
    \\\\
    \textbf{Baseline answer:} 

    In an isosceles right triangle, the altitude to the hypotenuse splits the triangle into two smaller isosceles right triangles and also bisects the hypotenuse. Let the legs of the original triangle be of length $x$. The hypotenuse, then, is $x\sqrt{2}$. \textcolor{red}{The altitude to the hypotenuse in an isosceles right triangle is also the geometric mean of the segments it divides the hypotenuse into,} which are each $\frac{x\sqrt{2}}{2}$. Therefore, the altitude $h$ can be expressed as:
\[ h = \frac{x\sqrt{2}}{2} \cdot \frac{x\sqrt{2}}{2} = \frac{x^2}{2} \]
Given that the altitude is $4\sqrt{2}$, we have:
\[ \frac{x^2}{2} = 4\sqrt{2} \]
Multiplying both sides by 2, we get:
\[ x^2 = 8\sqrt{2} \]
Taking the square root of both sides, we find:
\[ x = \sqrt{8\sqrt{2}} = 2\sqrt{2\sqrt{2}} \]
The area $A$ of the isosceles right triangle is given by:
\[ A = \frac{1}{2}x^2 = \frac{1}{2} \cdot 8\sqrt{2} = 4\sqrt{2} \]
Therefore, the area of the triangle is $\boxed{16}$.
Final Answer: The final answer is 16. I hope it is correct.
    \\\\
    \textbf{Our answer:} 
    
    Let the legs of the isosceles right triangle be $x$. Then, the hypotenuse is $x\sqrt{2}$. The altitude to the hypotenuse is $\frac{x\sqrt{2}}{2}$, \textcolor{red}{since the altitude of an isosceles right triangle is half the length of the hypotenuse.} We are given that the altitude is $4\sqrt{2}$, so we have

$$\frac{x\sqrt{2}}{2}=4\sqrt{2}\Rightarrow x\sqrt{2}=8\sqrt{2}\Rightarrow x=8.$$The area of the triangle is $\frac{1}{2}x^2=\frac{1}{2}(8^2)=\boxed{32}.$
Final Answer: The final answer is $32$. I hope it is correct.

\end{tcolorbox}

\end{document}